\documentclass[runningheads]{llncs}
\usepackage{graphicx}
\usepackage{subfigure}
\usepackage{color}
\usepackage{array}
\begin{document}
%
\title{Perceptual Compressive Sensing}
%
%
\author{Jiang Du \and
Xuemei Xie\thanks{Corresponding author: Xuemei Xie} \and
Chenye Wang \and
Guangming Shi}
\authorrunning{J. Du et al.}
%
\institute{School of Artificial Intelligence, Xidian University, Xi'an 710071, China\\
\email{jiangdu@ieee.org, xmxie@mail.xidian.edu.cn, cywang\_dd@163.com, gmshi@xidian.edu.cn}}
\maketitle              
\begin{abstract}
Compressive sensing (CS) works to acquire measurements at sub-Nyquist rate and recover the scene images. Existing CS methods always recover the scene images in pixel level. This causes the smoothness of recovered images and lack of structure information, especially at a low measurement rate. To overcome this drawback, in this paper, we propose perceptual CS to obtain high-level structured recovery. Our task no longer focuses on pixel level. Instead, we work to make a better visual effect. In detail, we employ perceptual loss, defined on feature level, to enhance the structure information of the recovered images. Experiments show that our method achieves better visual results with stronger structure information than existing CS methods at the same measurement rate.

\end{abstract}
\section{Introduction}

Nowadays, information is one of the most important component in human world. Visual information takes up most of the percentage. There are billions of images and videos around our daily life. Computer vision has underwent huge resurgence in recent years, since deep learning has made a significant difference in this field.
Researchers have shown that deep learning has made breakthrough achievements in the following two broad categories. The first category is the high-level computer vision tasks. For example, image and video classification or recognition~\cite{8253595}~\cite{Wang_2017_CVPR}~\cite{Fu_2017_CVPR}, object detection~\cite{He_2017_ICCV}~\cite{Lin_2017_ICCV}, image caption~\cite{Gan_2017_CVPR}~\cite{Yao_2017_ICCV}, and visual tracking~\cite{Song_2017_ICCV}~\cite{Yun_2017_CVPR}.  The second category is low-level reconstruction tasks. For example, image denoising, super-resolution~\cite{Dong2014Learning}~\cite{Ledig_2017_CVPR}, style transfer~\cite{Johnson2016Perceptual}~\cite{Chen_2017_CVPR}, and optical flow estimation~\cite{Ilg_2017_CVPR}~\cite{Cheng_2017_ICCV}.

Researches on inverse problems in imaging~\cite{Lucas2018Using}~\cite{mccann2017convolutional} have been carried on for decades, which cover various low-level computer vision tasks.
Compressive sensing (CS)~\cite{donoho2006cs}~\cite{Baraniuk2007Compressive}~\cite{Baraniuk2011More} is a typical inverse problem in imaging. Conventional CS works to recover the signal by optimization algorithms~\cite{Candes2005Decoding}~\cite{Figueiredo2008Gradient}.
However, this model is hard to be implemented and costs much computational complexity.
The application of deep neural networks in inverse problems in imaging makes it possible that the CS measurements can be recovered real-time.
Data-driven CS~\cite{Mousavi2016A}~\cite{Kulkarni2016ReconNet}~\cite{Mousavi2017Learning} learns the recovery network by the training data. Adp-Rec~\cite{Xie2017Adaptive} jointly train the coder-decoder and brings significant improvement on reconstruction quality. Fully convolutional measurement network (FCMN)~\cite{xie2017Fully} firstly measures and recovers full images.
However, all the above methods focus on pixel level, and ignore the high-level structure information. This makes the reconstructed results look smooth and have unsatisfactory visual effect.
To overcome the drawback, we consider to add high-level  perceptual information to CS. So the question is, how to add high-level perceptual information on the low-level CS task.

Recently, perceptual loss~\cite{Johnson2016Perceptual} has been used in many reconstruction tasks, such as style transfer~\cite{Johnson2016Perceptual}~\cite{Chen_2017_CVPR}.
They are a combination of low-level detailed information and high level semantic information. Perceptual loss is a widely used way to achieve these goals. It is because perceptual loss is defined in feature space, which can convert the ability of catching high-level structure information to recovery network. Thus, the recovered images will contain rich structure information. Inspired by the above applications, we propose perceptual CS, which focuses more on sensing and recovering structure information. Here perceptual loss is employed on CS framework. We use FCMN \cite{xie2017Fully} as base network to measure and recover scene images, and adopt perceptual loss to train it. We surprisingly find that this framework is capable of capturing and recovering the structure information, especially at extremely low measurement rate, where the measurements can merely contain very limited amount of information.

The contribution of this paper is that,
we propose perceptual CS, which can measure and recover the structure information of scene images.
It should be pointed out that, only one deconvolutional layer and one Res-block are used in our proposed framework. This is just an illustration.
One can employ a deeper network architecture if necessary.

Moreover, perceptual CS indicates an universal architecture. One can change the loss network using pre-trained or dynamic feature extractors for more specific tasks.
In this paper, we use VGG~\cite{Simonyan2014Very} as an example. Our code will be available on github\footnote{https://github.com/jiang-du/Perceptual-CS} for further reproduction.

The organization of the rest part of this paper is as follows. Section 2 introduces some related works of this paper. Section 3 describes the technical design and theoretical analysis of the proposed framework. Section 4 presents experimental results of perceptual CS and gives detailed analysis. Section 5 draws the conclusion.

\section{Related work}

\subsection{Compressive Sensing}

CS \cite{donoho2006cs} \cite{Kunis2008Random} \cite{Tropp2007Signal} proves signal can be reconstructed after being sampled at sub-Nyquist rates as long as the signal is sparse in a certain domain. Reconstructing signal from measurements is an ill-posed problem. Traditional CS usually solves an optimization problem, which leads to high computational complexity.

Recently, deep neural networks (DNNs) has been applied to CS tasks \cite{Mousavi2016A} \cite{Kulkarni2016ReconNet} \cite{Mousavi2017Learning} \cite{Mousavi2017DeepCodec} \cite{Xie2017Adaptive} \cite{xie2017Fully}. These DNN-based methods methods can be divided into two categories depending on whether measurement and reconstruction process are trained jointly.
The first category trains the recovery network while the measurement part is fixed, like SDA \cite{Mousavi2016A}, ReconNet \cite{Kulkarni2016ReconNet}, and DeepInverse \cite{Mousavi2017Learning}.
SDA \cite{Mousavi2016A} first applies deep learning approach to solve the CS recovery problem, which uses fully-connected layers in the recovery part.
ReconNet~\cite{Kulkarni2016ReconNet} uses a fully-connected layer along with convolutional layers to recover signals block by block.
While, DeepInverse \cite{Mousavi2017Learning} uses pure convolutional layers.
The random Gaussian fashion of the measurement part will mismatch the learned recovery part.

The second category jointly trains the measurement part and the recovery part,
such as Deepcodec \cite{Mousavi2017DeepCodec}, Adaptive \cite{Xie2017Adaptive}, and FCMN \cite{xie2017Fully}.
These methods totally overcome the problem that the measurement part is independent from the recovery part.
Deepcodec \cite{Mousavi2017DeepCodec} is a framework where both measurement and approximate inverse process are learned end-to-end by a deep fully-connected encoder-decoder network. In \cite{Xie2017Adaptive}, a fully-connected layer as the measurement matrix along with a super-resolution network as the recovery part is trained. FCMN \cite{xie2017Fully} firstly uses a fully convolutional network where the measurement part is implemented with an overlapped convolution operation.
All these methods recover the scene image on pixel level. They ignore the structure information of images.

\subsection{Perceptual Loss}

Recently, perceptual loss \cite{Johnson2016Perceptual} is widely used in many image reconstruction tasks \cite{Johnson2016Perceptual} \cite{Isola_2017_CVPR} \cite{Yeh2017Semantic} \cite{Ledig_2017_CVPR} \cite{Shen_2017_CVPR} \cite{Huang2017Beyond}.
It can recover the image with better visual effect since it is defined on feature space. Typically, perceptual loss calculates the Euclidean distance between the features maps of the reconstructed images and the labels from the same layer of the same pre-trained classification network.
Perceptual loss reflects the similarity in the feature level between the label and output images, which makes the reconstructed images retain high-level structure information. In contrast, per-pixel loss focuses on similarity in pixel level, which only preserves low-level pixel information.

Perceptual loss achieves more excellent performance than per-pixel loss in most of image restoration tasks. For example, Johnson et al. \cite{Johnson2016Perceptual} use perceptual loss for style transfer and super resolution. The output images have sharper edges compared to per-pixel loss. SRGAN \cite{Ledig_2017_CVPR} trained by perceptual loss generates more photo-realistic super-resolved images than by MSE loss. When used in image inpainting \cite{Shen_2017_CVPR}, perceptual loss produces satisfactory results due to the addition of high-level context. Additionally, perceptual loss helps to remain finer details for image editing \cite{Yeh2017Semantic}. Inspired by the advantages of perceptual loss in preserving structure and detail, we attempt to apply it to CS field and it accordingly performs well.

\section{Perceptual CS Framework}

\begin{figure*}[!h]
  \centering
  \includegraphics[width=\textwidth]{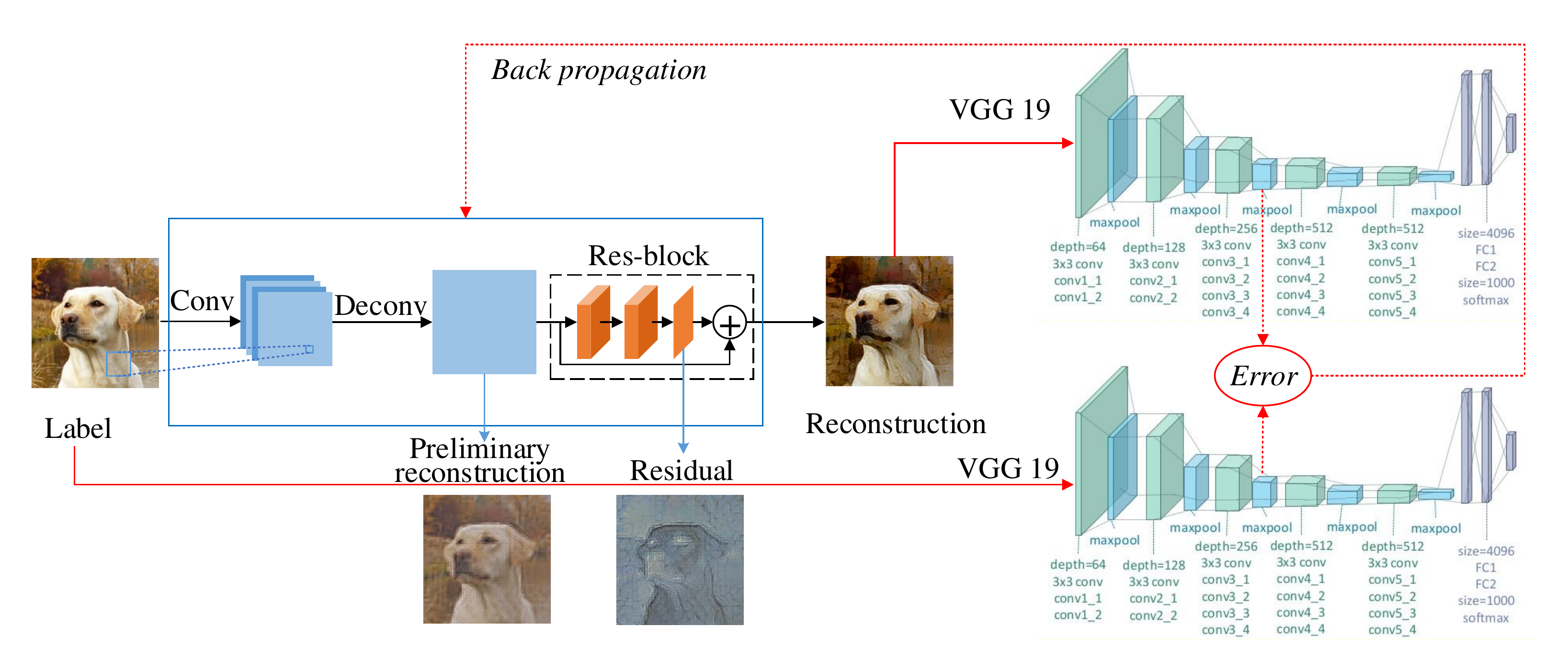}\\
  \caption{The architecture of perceptual CS network.}
  \label{fig:framework}
\end{figure*}

In this section, we mainly introduce the technical design of the perceptual CS framework. The architecture is shown in Fig. \ref{fig:framework}. It consists of two parts: \emph{compressive sensing network} and \emph{perceptual loss network}.
The compressive sensing network originally performs reconstruction in pixel-wise manner. 
With the perceptual loss network added, the perceptual CS network preserves the structure information of the recovered images.
With the help of perceptual recovery, the proposed network is able to acquire high-level perceptual information.

The compressive sensing network measures and recovers the full scene images.
The full image processing fashion provides an enough receptive field that makes it possible to perform perceptual reconstruction.
While, in the perceptual loss network, we employ a classification network, VGG19, as an auxiliary network. It plays the role of extracting the perceptual information of the images.

\subsection{Full Image Compressive Sensing Network}

In most existing CS methods, the scene image is measured and recovered block by block, and each block is reshaped into a column vector.
This breaks the structure of the full image. Besides, the computational complexity of the existing methods will extremely increase when the size of the image becomes larger.
For example, when an image with the size of $n\times n$ is measured, the memory consumption of the sensing matrix can be up to $S(n)=O(n^4)$.
Thus, it is nearly impossible to design a large sensing matrix, let alone measuring the full image. 
This is because the mapping from the scene image to the measurements is fully-connected, leading to an extremely large-scale parameter nightmare.

Inspired by fully convolutional measurement network (FCMN)~\cite{xie2017Fully},
we employ a fully convolutional architecture to measure and recover the scene images in the proposed framework,
which can get rid of the disaster of the exploding number of parameters.
Moreover, the fully convolutional architecture can preserve the correspondence among pixels (instead of reshaping into column vector).
In this way, block-effect has been largely removed in the recovered images due to the overlapped convolutional measurement.
This preserves the structure information of the whole image. Furthermore, the full image method makes it possible to use perceptual loss for semantic reconstruction.

\begin{figure}[!tp]
  \centering
  \subfigure[Original]{
    \begin{minipage}[b]{0.22\textwidth}
    \includegraphics[width=1\textwidth]{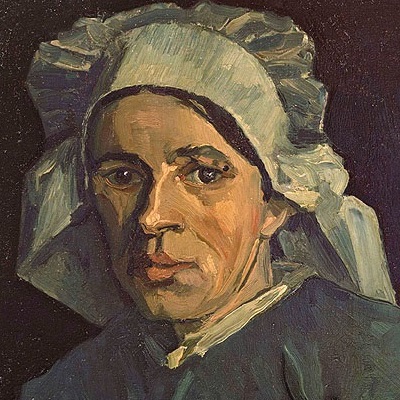}
    \end{minipage}
  }
  \subfigure[\newline FCMN~\cite{xie2017Fully}]{
    \begin{minipage}[b]{0.22\textwidth}
    \includegraphics[width=1\textwidth]{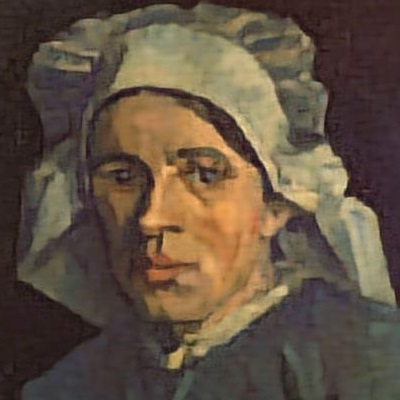}
    \end{minipage}
  }
  \subfigure[VGG$_{2\_2}$]{
    \begin{minipage}[b]{0.22\textwidth}
    \includegraphics[width=1\textwidth]{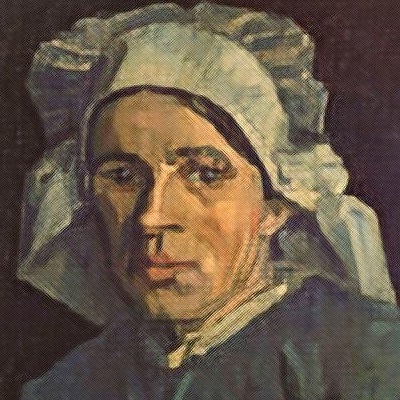}
    \end{minipage}
  }
  \subfigure[VGG$_{3\_4}$]{
    \begin{minipage}[b]{0.22\textwidth}
    \includegraphics[width=1\textwidth]{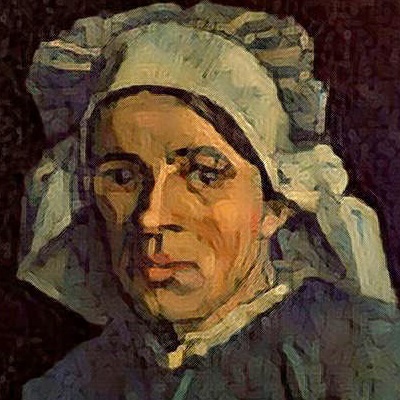}
    \end{minipage}
  }
  \caption{The original image `Head of a peasant woman with white cap' by Van Gogh and the reconstructed images with different methods at 4\% measurement rate. Here the proposed method uses the conv2\_2 and conv3\_4 of VGG19~\cite{Simonyan2014Very} as different scale of loss respectively.}
  \label{fig:vangogh}
\end{figure}

Although the convolution and deconvolution layers can recover the image, for better visual effect, we enhance the proposed framework with residual learning. 
In detail, we just add one residual block and it works quite well, as is shown in Fig \ref{fig:vangogh} (b).
One can add more residual blocks for further improvements if necessary. 

\subsection{Perceptual Reconstruction for Compressive Sensing}

In the proposed network, we focus on the perceptual recovery.
In the classic CS task, the recovery network approximates the error in the pixel-wise space.
To extract the structure information, we recover the scene image in feature-level space.
Instead of MSE loss, we consider the perceptual loss, which focuses on perceptual recovery.

\emph{MSE loss}: In classic CNN-based CS, the loss function is usually defined with pixel-wise loss:
\begin{equation}
  l_{pixel}(w)=\left \| f\{x,w\}-x \right \|_2^2.
\end{equation}
This pixel-wise loss will force the image to have the minimized average Euclidean distance between the reconstruction images $f\{x,w\}$ and the labels $x$.
Here, $w$ represents the parameters of the whole network, including the measurement and the recovery parts.
Although MSE loss in (1) can help to achieve the reconstructed images with high peak signal-to-noise ratio (PSNR), the reconstructed images usually look smooth and the structure information is not clear.
We can see in Fig \ref{fig:vangogh} (b) that the face and the hat of the person is very smooth compared with the original image in Fig \ref{fig:vangogh} (a). Especially the wrinkle on the face cannot be clearly seen.

\emph{Perceptual loss}: Considering the current popular classification network works by extracting the features in an image, we can take this advantage into our proposed method. Thus, we apply the perceptual loss.
It is formulated as
\begin{equation}
  l_{feat}^{\phi,j}(w)=\left \| \phi_j(f\{x,w\})-\phi_j(x) \right \|_2^2,
\end{equation}
where $\phi$$_j(x)$ denotes the feature map of the $j$-th layer of VGG19 with the input image $x$.
Different from (1), a typical kind of perceptual loss is defined with the (squared, normalized) Euclidean distance 
between the feature maps generated from the reconstructed image and the label.
Actually, when applying CS at a very low measurement rate, we do not care much about the detailed texture of it.
Correspondingly, we emphasize the importance of the structural information. As is shown in Fig \ref{fig:vangogh} (c) and (d), the structure information recovered better, especially the hat of the person has richer structure information compared with Fig \ref{fig:vangogh} (b).

In practical, we define the loss function on VGG$_{2\_2}$ or VGG$_{3\_4}$ of VGG19 (actually pooling 2 or pooling 3) as examples. The results can be addressed in Fig \ref{fig:vangogh} (c) and (d). The feature map of bottom layers contains detailed low-level information and the top layers have more high-level semantic features. We can also choose other layers by different requirements. In this paper, We do not apply perceptual loss by too high level layers because in terms of compressive sensing, higher level drops too much information that it is nearly impossible to inverse, even if pre-trained.

\section{Experiments with Analysis}

In this section, we conduct the experiments to illustrate the performance of the proposed perceptual CS framework. We test our framework with a standard dataset \cite{Kulkarni2016ReconNet} containing 11 grayscale images.
We also compare the reconstruction results with some typical CS methods. Furthermore, we take some reconstruction results as examples to make a detailed analysis of the performance of the proposed method.

{\bf Experiment Setup}
The learning rate is set to $10^{-8}$ when perceptual loss is defined on VGG$_{2\_2}$, and $10^{-9}$ when perceptual loss is defined on VGG$_{3\_4}$. The bench size is set to 5 while training. For each measurement rate, the iteration time is $10^{6}$. We use the caffe~\cite{Jia2014Caffe} framework for network training and MATLAB for testing. Our computer is equipped with Intel Core i7-6700K CPU with frequency of 4.0GHz, 4 NVidia GeForce GTX Titan XP GPUs, 128 GB RAM, and the framework runs on the Ubuntu 16.04 operating system. The training dataset consists of 800 pieces of $256\times 256$ size images down sampled and cropped from 800 images in DIV2K dataset~\cite{agustsson2017ntire}.

{\bf Results with analysis} The following is the analysis of the experimental results at different measurement rates.

\begin{figure}[!h]
  \centering
  \subfigure[Original]{
    \begin{minipage}[b]{0.3\textwidth}
    \centering
    \includegraphics[width=0.8\textwidth]{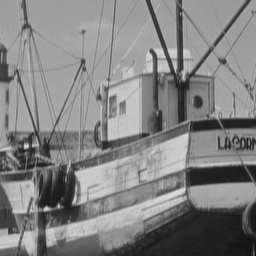}
    \end{minipage}
  }
  \subfigure[ReconNet(18.93dB)]{
    \begin{minipage}[b]{0.3\textwidth}
    \centering
    \includegraphics[width=0.8\textwidth]{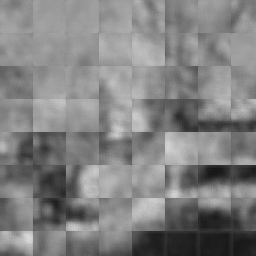}
    \end{minipage}
  }
  \subfigure[Adp-Rec(21.67dB)]{
    \begin{minipage}[b]{0.3\textwidth}
    \centering
    \includegraphics[width=0.8\textwidth]{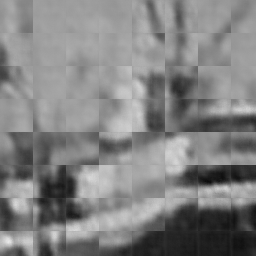}
    \end{minipage}
  }
  \subfigure[FCMN(22.49dB)]{
    \begin{minipage}[b]{0.3\textwidth}
    \centering
    \includegraphics[width=0.8\textwidth]{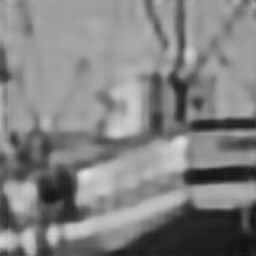}
    \end{minipage}
  }
  \subfigure[\quad Proposed VGG$_{2\_2}$\newline \centerline{(19.38dB)}]{
    \begin{minipage}[b]{0.3\textwidth}
    \centering
    \includegraphics[width=0.8\textwidth]{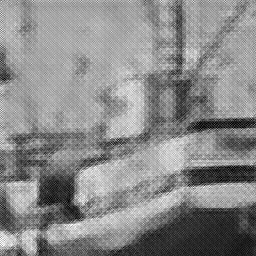}
    \end{minipage}
  }
  \subfigure[\quad Proposed VGG$_{3\_4}$\newline \centerline{(18.07dB)}]{
    \begin{minipage}[b]{0.3\textwidth}
    \centering
    \includegraphics[width=0.8\textwidth]{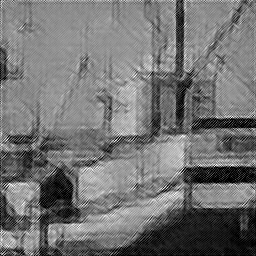}
    \end{minipage}
  }
  \caption{Boats at measurement rate 1\%. (b) and (c) are of block-wise. (d) is of full-image. They all use MSE loss. (e) and (f) are improved by using perceptual loss [2]. Perceptual CS brings stronger structure information compared with FCMN.}
  \label{fig:boats}
\end{figure}

\begin{figure}[!h]
  \centering
  \subfigure[Original]{
    \begin{minipage}[b]{0.3\textwidth}
    \centering
    \includegraphics[width=0.8\textwidth]{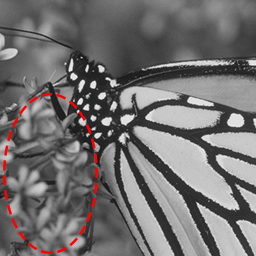}
    \end{minipage}
  }
  \subfigure[ReconNet(18.19dB)]{
    \begin{minipage}[b]{0.3\textwidth}
    \centering
    \includegraphics[width=0.8\textwidth]{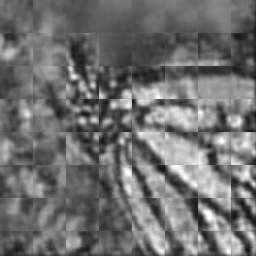}
    \end{minipage}
  }
  \subfigure[DR$^2$-Net(18.93dB)]{
    \begin{minipage}[b]{0.3\textwidth}
    \centering
    \includegraphics[width=0.8\textwidth]{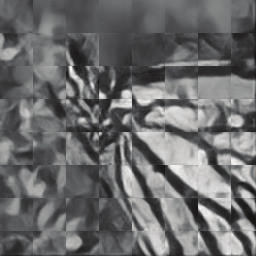}
    \end{minipage}
  }
  \subfigure[FCMN(22.52dB)]{
    \begin{minipage}[b]{0.3\textwidth}
    \centering
    \includegraphics[width=0.8\textwidth]{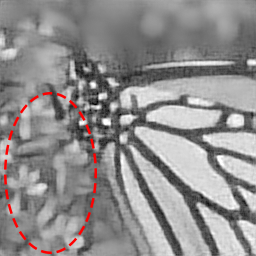}
    \end{minipage}
  }
  \subfigure[\quad Proposed VGG$_{2\_2}$\newline \centerline{(18.25dB)}]{
    \begin{minipage}[b]{0.3\textwidth}
    \centering
    \includegraphics[width=0.8\textwidth]{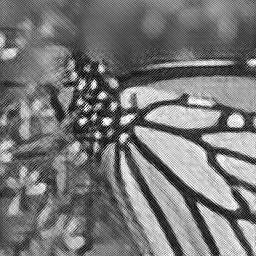}
    \end{minipage}
  }
  \subfigure[\quad Proposed VGG$_{3\_4}$\newline \centerline{(16.35dB)}]{
    \begin{minipage}[b]{0.3\textwidth}
    \centering
    \includegraphics[width=0.8\textwidth]{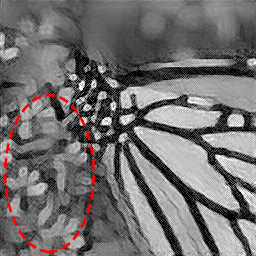}
    \end{minipage}
  }
  \caption{Monarch at measurement rate 4\%. (b) and (c) are of block-wise. (d) is of full-image. They all use MSE loss. (e) and (f) are improved with perceptual loss. They have stronger structure information than the state-of-the-art result in FCMN. Specially, we can see in the \textcolor[rgb]{1.00,0.00,0.00}{red circle} of (f), compared with (a) and (d), that even blurry image can be enhanced.}
  \label{fig:Monarch}
\end{figure}

The explanation from Fig \ref{fig:boats} at measurement rate 1\% is as follows.

(1)	Block effect occurs in Fig \ref{fig:boats} (b) and (c) by block-wise methods such as ReconNet \cite{Kulkarni2016ReconNet} and Adp-Rec \cite{Xie2017Adaptive}.

Based on the standard ReconNet \cite{Kulkarni2016ReconNet}, the improved ReconNet \cite{Lohit2017Convolutional} adds several tricks such as adaptive measurement and adversarial loss. Its performance is even lower than Adp-Rec \cite{Xie2017Adaptive}.

(2)	Fig \ref{fig:boats} (d) has no block artifacts in FCMN \cite{xie2017Fully} where fully-convolutional measurement is employed. This work achieves the state-of-the-art results in terms of PSNR and SSIM.

All existing CS-based image reconstruction works rely on MSE loss. While, FCMN \cite{xie2017Fully} makes perceptual loss promising.

(3)	Perceptual loss in Fig \ref{fig:boats} (e) and (f) enhances structure information, even if PSNR is lower compared with Fig \ref{fig:boats} (d).
\newline

The explanation of measurement rate=4\% in Fig \ref{fig:Monarch} is as follows:

(1)	Block effect also occurs in Fig \ref{fig:Monarch} (c) in DR$^2$-Net \cite{Yao2017DR2}.

DR$^2$-Net achieves highest PSNR among random Gaussian methods, since it adds several Res-blocks that fully convergence in the reconstruction stage.

(2)	The method with adaptive measurement for Fig \ref{fig:Monarch} (d) adopts one Res-block, achieving the highest PSNR. The comparison among several typical methods including DR$^2$-Net is in Fig \ref{fig:Monarch}, where FCMN \cite{xie2017Fully} with full image gets the best result in terms of PSNR.

It should be pointed out that only one Res-block is used in both FCMN \cite{xie2017Fully} and the proposed framework in this paper. One can add more Res-blocks for further improvement.

(3)	With just one Res-block, perceptual loss in Fig \ref{fig:Monarch} (e) and (f) works well, which improves FCMN \cite{xie2017Fully}. Structure information is kept. In some case, even wake structure can become strong (see Fig \ref{fig:Monarch} (f) compared to Fig \ref{fig:Monarch} (a) and (d)).

It should be noted that, even if PSNR is worse with perceptual loss, the structure information is clearly reconstructed.

\setlength{\tabcolsep}{4pt}
\begin{table}[!b]
\begin{center}
\caption{Mean PSNR, SSIM and MOS of different methods}
\label{table:PSNR}
\begin{tabular}{p{1.5cm}p{1.4cm}<{\centering}p{1.4cm}<{\centering}p{1.4cm}<{\centering}p{1.4cm}<{\centering}p{1.4cm}<{\centering}p{1.4cm}<{\centering} }
\textbf{MR=1\%} & ReconNet & DR$^2$-Net & Adp-Rec & FCMN & VGG$_{2\_2}$ & VGG$_{3\_4}$\\
\noalign{\smallskip}
\hline
\noalign{\smallskip}
PSNR        & 17.27 & 17.44 & 20.32 & \textbf{21.27} & 18.30 & 16.80\\
SSIM     & 0.4083 & 0.4291 & 0.5031 & \textbf{0.5447} & 0.2478& 0.2565\\
MOS        & 1.0734 & 1.1188 & 1.8496 & 2.6328 & 2.6818& \textbf{\textcolor[rgb]{1.00,0.00,0.00}{2.9510}}\\
         \\
\textbf{MR=4\%}\\
\noalign{\smallskip}
\hline
\noalign{\smallskip}
PSNR        & 19.99 & 20.80 & \textbf{24.01} & 23.87 & 19.38  & 16.72\\
SSIM     & 0.5287 & 0.5804 & 0.7021 & \textbf{0.7042} & 0.3522 & 0.4729\\
MOS        & 1.5979 & 1.7237 & 3.0489 & 3.4230 & \textbf{\textcolor[rgb]{1.00,0.00,0.00}{3.4755}} & 3.3566\\
\end{tabular}
\end{center}
\end{table}
\setlength{\tabcolsep}{1.4pt}

{\bf Evaluation of Perceptual CS.}  To evaluate the performance of the proposed method, we evaluate quality of the reconstructed images with PSNR and SSIM.
Furthermore, we also use Mean Opinion Score (MOS)~\cite{recommendatios2000recommendation} to test the visual effect of these methods.
In this metric, an image is scored by 26 volunteers and the final score is the average value.
The quality ranking is represented by scores from 1 to 5, where 1 denotes lowest quality and 5 denotes the highest.
All the test images are ranked randomly before being scored and they are displayed group by group. Each group has six reconstruction images, in different methods. All participants take this test on the same computer screen, from the same angle and distance. Here the distance from the screen to the tested persons is 50 cm and the eyes of those persons are of the same height of the center of the screen.

\begin{figure*}[!b]
  \centering
  \includegraphics[width=\textwidth]{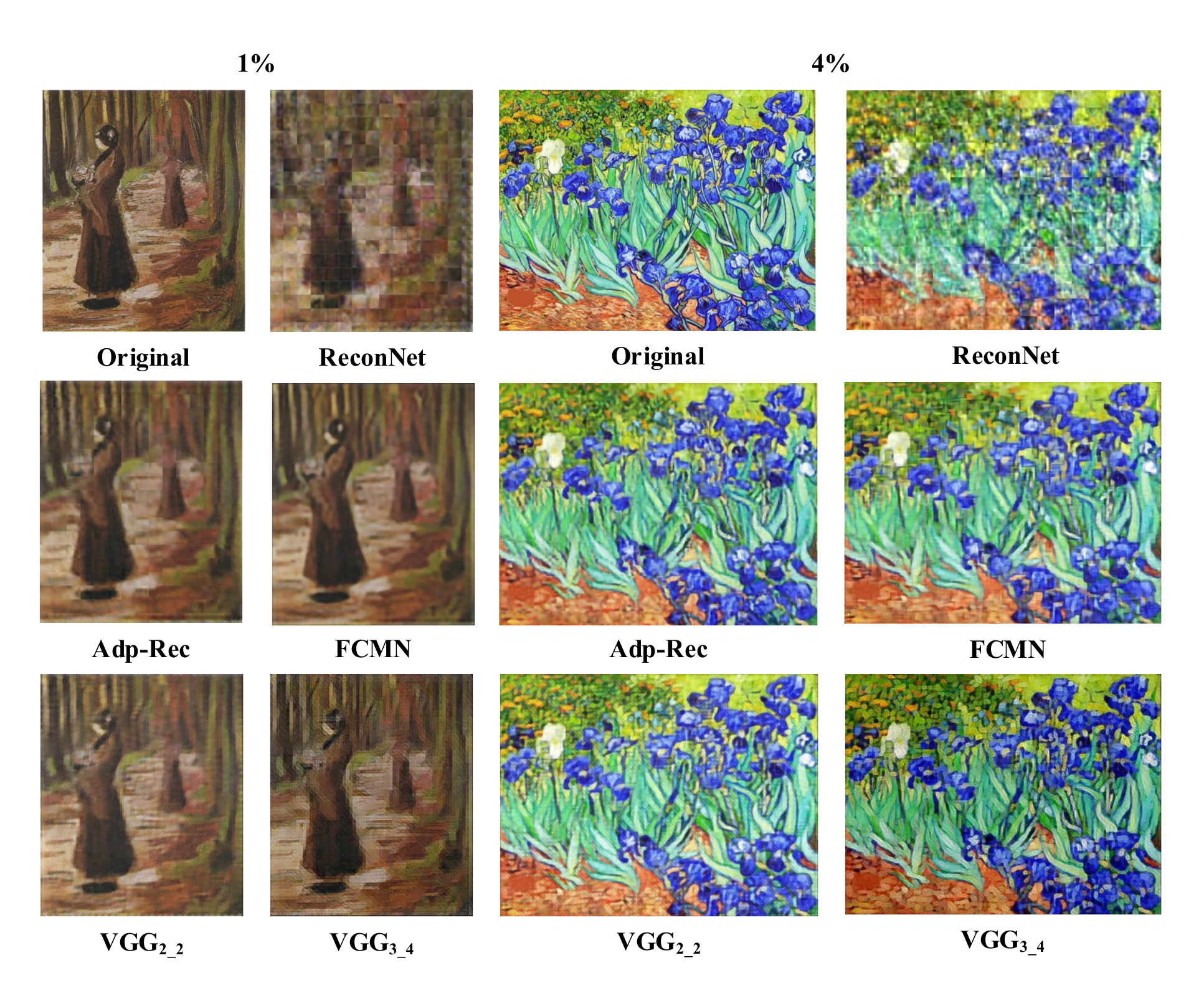}\\
  \caption{The reconstructed results of ReconNet~\cite{Kulkarni2016ReconNet}, Adp-Rec~\cite{Xie2017Adaptive}, FCMN~\cite{xie2017Fully}, and the proposed method using the conv2\_2 and conv3\_4 of VGG19~\cite{Simonyan2014Very} with measurement rate 1\% and 4\% and their corresponding original scene image.}
  \label{fig:fangao}
\end{figure*}

The detailed comparison results of mean PSNR, SSIM and MOS is shown in Table \ref{table:PSNR}. we can draw the following conclusion. Our method achieves the highest MOS rating. The PSNR and SSIM value of typical methods is higher, since their loss function is defined as the Euclidean distance between the output and label. While, perceptual CS concentrates more on the visual effect. Thus, it is helpful for MOS, instead of PSNR and SSIM.

Moreover, we give some examples of color images. In terms of color channels, we measure and recover the RGB channels respectively, and then combine them to a whole color image. The results of perceptual CS with color images are shown in Fig \ref{fig:fangao}. Of course, we give the comparison with existing methods. We can see obviously from the figure that the visual effect of perceptual CS is quite well.

In terms of hardware implementation, we follow the approach of the existing work proposed in \cite{shi2011hr} in which sliding window is used to measure the scene. Similarly, we can replace the random Gaussian measurement matrix with the learned pre-defined parameters in the convolution layer of the measurement network. The reconstruction part is not on optical device, so only the measurement part needs to be implemented with the approach above.

\section{Conclusion}

In this paper, we propose perceptual CS for sensing and recovering structured scene images.
The proposed framework managed to recover structure information from CS measurements.
Our work is of profound significance, which may open a door towards alternative to semantic sensing and recovery.

\section*{Acknowledgements}

This work is supported by Natural Science Foundation (NSF) of China (61472301, 61632019) and Ministry of Education project (6141A02011601).

%
%
%
 \bibliographystyle{splncs04}
 \bibliography{PRCV65}

\end{document}